\documentclass[letterpaper, 10 pt, conference]{ieeeconf} 

\IEEEoverridecommandlockouts                              %
\usepackage{graphics} %
\usepackage{epsfig} %
\usepackage{mathptmx} %
\usepackage{times} %
\usepackage{amsmath} %
\usepackage{amssymb}  %
\usepackage{url}
\usepackage{booktabs} 
\usepackage{array}  
\usepackage{multirow} 
\usepackage{makecell}

\usepackage{colortbl}
\usepackage{xcolor}
\usepackage{framed}
\usepackage{wrapfig}  
\usepackage{titletoc}
\usepackage{color} 
\usepackage{caption, subcaption, overpic, textpos}
\usepackage{titlesec}
\usepackage{comment}
\usepackage{fontawesome}

\usepackage{tcolorbox} %
\tcbuselibrary{listings} %

\usepackage{xspace}
\newcommand{\projname}{AgiBot World\xspace}
\newcommand{\modelname}{GO-1\xspace}

\newcommand{\embodiment}{AgiBot G1\xspace}

\usepackage{pifont}%
\newcommand{\cmark}{\ding{51}}%
\newcommand{\xmark}{\ding{55}}%

\definecolor{deemph}{gray}{0.6}

\definecolor{baselinecolor}{gray}{.9}
\definecolor{yellow}{RGB}{218,165,32}
\definecolor{lightcyan}{rgb}{0.88, 1.0, 1.0}
\definecolor{lightskyblue}{rgb}{0.53, 0.81, 0.98}
\definecolor{aliceblue}{rgb}{0.94, 0.97, 1.0}
\definecolor{LightSlateBlue}{RGB}{70,130,180}
\definecolor{DeepBlue}{RGB}{65,100,170}
\definecolor{DeepPurple}{RGB}{136,105,160}
\definecolor{LightGreen}{RGB}{59,125,35}
\definecolor{LightRed}{RGB}{234,66,53}
\definecolor{cvprblue}{rgb}{0.21,0.49,0.74}

\definecolor{mypink}{RGB}{254,102,140}
\definecolor{myclay}{RGB}{59,182,176}

\newtcblisting{bibtexcode}{
  colback=gray!5, %
  colframe=white, %
  boxrule=0pt, %
  arc=0pt, %
  left=6pt, %
  right=6pt, %
  top=6pt, %
  bottom=6pt, %
  listing only, %
  listing options={
    basicstyle=\ttfamily\footnotesize, %
    breaklines=true, %
    language={} %
  }
}

\makeatletter
\let\NAT@parse\undefined
\makeatother
\usepackage[pagebackref=true,breaklinks,colorlinks,citecolor=cvprblue,bookmarks=true]{hyperref}\usepackage[capitalize]{cleveref}

\crefname{section}{Sec.}{Secs.}
\crefname{section}{Sec.}{Secs.}

\crefname{figure}{Fig.}{Figs.} %
\Crefname{figure}{Fig.}{Figs.} %

\crefname{table}{Tab.}{Tabs.} %
\Crefname{table}{Tab.}{Tabs.} %

\title{\LARGE \bf %
 \includegraphics[height=6mm]{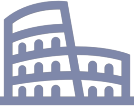} AgiBot World Colosseo: A Large-scale Manipulation Platform \\ for 
Scalable and Intelligent Embodied Systems
}

\author{Team AgiBot-World$^*$ \\
{\small Project website: \url{https://agibot-world.com/}} \\
{\small Code: \url{https://github.com/OpenDriveLab/AgiBot-World}}
}

\begin{document}

\pagestyle{empty} 

\noindent
\twocolumn[{
\renewcommand\twocolumn[1][]{#1}

\maketitle

\vspace{-5mm}
\begin{center}
    \centering
    \captionsetup{type=figure}
    \includegraphics[width=\textwidth]{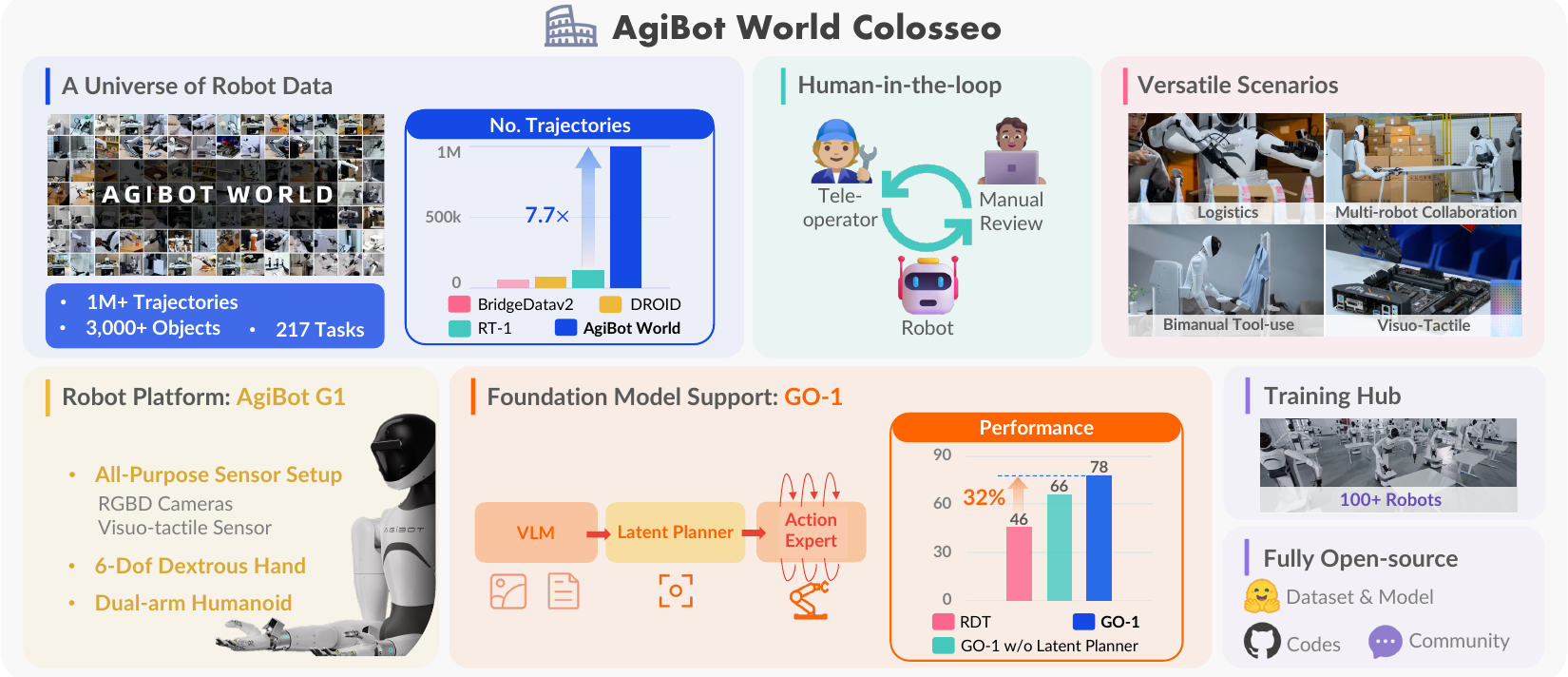}
    \captionof{figure}{Introducing \textbf{AgiBot World Colosseo},  an open-sourced large-scale manipulation platform comprising data, models, benchmarks and ecosystem. \projname stands out for its unparalleled {scale and diversity} compared to prior counterparts.
    A suite of 100 dual-arm humanoid robots
    is deployed.
    We further propose a generalist policy (\modelname) with the latent action planner. 
    It is trained 
    across diverse
    data corpus with a
    scalable performance  
    of {32\%} gain compared to prior arts.
    \label{fig:teaser}
    }
\end{center}

}]

\begin{abstract}
We explore how scalable robot data can address real-world challenges for generalized robotic manipulation. Introducing AgiBot World, a large-scale platform comprising over 1 million trajectories across 217 tasks in five deployment scenarios, we achieve an order-of-magnitude increase in data scale compared to existing datasets.
Accelerated 
by a standardized collection pipeline with 
human-in-the-loop verification, AgiBot World guarantees high-quality and diverse data distribution.
It is extensible from grippers to dexterous hands and visuo-tactile sensors for fine-grained skill acquisition.
Building on top of data, we introduce Genie Operator-1 (\modelname), a novel generalist policy that leverages latent action representations to maximize data utilization, demonstrating predictable performance scaling with increased data volume.
Policies pre-trained on our dataset achieve an average performance improvement of 
30\% over those trained on Open X-Embodiment, both in in-domain and out-of-distribution scenarios.
\modelname 
exhibits
exceptional capability in real-world dexterous and long-horizon tasks, achieving 
over 60\% success rate on complex tasks
and outperforming prior RDT approach by 32\%. 
By open-sourcing the dataset, tools, and models, we aim to democratize access to large-scale, high-quality robot data, advancing the pursuit of scalable and general-purpose intelligence.
\let\thefootnote\relax\footnote{
\hspace{-0.05\columnwidth}
$^*$ This work is a joint effort among The University of Hong Kong, 
AgiBot Inc., Shanghai Innovation Institute and 
Shanghai AI Lab. For detailed authorship, please consult our 
website and GitHub repository.
}

\end{abstract}

\section{Introduction}

Manipulation is a cornerstone task in robotics,
enabling the agent to interact with and adapt to the physical world. While significant progress has been made in general-purpose foundational models for natural language processing~\cite{achiam2023gpt} and computer vision~\cite{ravi2024sam}, robotics lags behind due to the difficulty of (high-quality) data collection. 
In the controlled lab setting, simple tasks such as pick-and-place have been well studied~\cite{chi2023diffusion,kim2024openvla}.
Yet for the open-set
real-world setting, tasks spanning from fine-grained object interaction, mobile manipulation to collaborative tasks, remains a formidable challenge~\cite{cui2021towards}. 
These tasks require not only physical dexterity but also the ability to generalize across diverse environment and scenarios, a 
merit
beyond the reach of 
current robotic systems. The widely accepted reason is the lack of high-quality data---unlike images and text, which are abundant and standardized, robotic datasets suffer from fragmented clips due to heterogeneous hardware and unstandardized collection procedure, leading to low-quality and inconsistent outcome. 
In this work we ask, \textit{how could we resolve the real-world complexity 
effectively 
by scaling up real-world robot data?} 

\medskip
Recent efforts, such as Open X-Embodiment (OXE)~\cite{padalkar2023open}, have 
addressed 
by aggregating and standardizing existing datasets. Despite advancements on large-scale cross-embodiment learning, the resulting policy is constrained within naive, short-horizon tasks and can weakly generalize to out-of-domain scenarios~\cite{kim2024openvla}. 
DROID~\cite{khazatsky2024droid} collected expert data through crowd-sourcing from diverse real-life scenes. The absence of data quality assurance (with human feedback) and the reliance on a constrained hardware setup (\textit{i.e.,} featuring fixed, single-arm robots), limit its real-world applicability and broader effectiveness. More recently, Lin \textit{et al.}~\cite{lin2024data} explored scaling laws governing generalizability across intra-category objects and environments, albeit limited to a few simple, single-step tasks.
These efforts represent a notable advancement toward developing generalist policies, moving beyond the traditional focus on single-task learning within narrow domains~\cite{zhao2023learning,chi2023diffusion}. Nevertheless, existing robot learning datasets remain constrained by their reliance on short-horizon tasks in highly controlled laboratory environments, failing to adequately capture the complexity and diversity inherent in real-world manipulation tasks.
To achieve general-purpose robotic intelligence, it is essential to develop datasets that scale in size and diversity while capturing real-world variability, supported by general-purpose humanoid robots for robust skill acquisition, a standardized data collection pipeline with assured quality, and carefully curated tasks reflecting real-world challenges.

As depicted in \Cref{fig:teaser}, we introduce \textbf{ \projname Colosseo}, a full-stack large-scale robot learning platform curated for advancing
bimanual manipulation in scalable and intelligent embodied systems. 
A 
full-scale
4000-square-meter facility is constructed to represent five major 
domains---domestic, retail, industrial, restaurant, and office environment---all dedicated to high-fidelity data collection in authentic everyday scenarios. With over 1 million trajectories collected from 100 real robots, \projname offers unprecedented diversity and complexity. It spans over 100 real-world scenarios, addressing challenging tasks such as fine-grained manipulation, tool usage, and 
multi-robot
synergistic collaboration. 
Unlike prior
datasets, \projname dataset collection is carried out with a fully standardized pipeline, ensuring high data quality and scalability, while incorporating human-in-the-loop verification to guarantee reliability. 
Our hardware setup includes mobile base humanoid robots with whole-body control, dexterous hands, and visuo-tactile sensors, enabling rich, multimodal data collection. Each episode is meticulously designed, featuring multiple camera views, depth information, camera calibration, and language annotations for both the overall task and each individual sub-steps. This well-rounded hardware setup, combined with various long-horizon, real-world tasks, opens new avenues for developing next-generation generalist policies and fosters diverse future research in robotics.

Our experimental results highlight the transformative potential of the \projname dataset. Policies pre-trained on our dataset achieve an average success rate improvement of 30\% compared to those trained on the prior large-scale robot dataset OXE~\cite{padalkar2023open}. Notably, even when utilizing only a fraction of our dataset---equivalent to 1/10 of the data volume in hours compared to OXE---the generalizability of pretrained policies is elevated by 18\%. These findings underscore the dataset's efficacy in bridging the gap between controlled laboratory environments and real-world robotic applications.
Following our dataset, to address the limitations of previous robot foundation models that heavily rely on in-domain robot datasets, we present Genie Operator-1 (\modelname), a novel generalist policy that utilizes latent action representations to enable learning from heterogeneous data and efficiently bridges general-purpose vision-language models (VLMs) with robotic sequential decision-making. Through unified pre-training on web-scale data, spanning human videos to our high-quality robot dataset, \modelname achieves superior generalization and dexterity, outperforming prior generalist policies such as RDT~\cite{liu2024rdt} and our variant without latent action planner.
Moreover, we demonstrate that \modelname's performance exhibits robust scalability with increasing dataset size, underscoring its potential for sustained advancement as larger datasets become available.
Beyond its immediate impact, \projname lays a strong foundation for future research in robotic manipulation. By open-sourcing the dataset, toolchain, and pre-trained models, we aim to foster community-wide innovation, enabling researchers to explore more authentic and diverse applications from household assistant to industrial automation. \projname is more than yet another dataset; it is a step toward scalable, general-purpose robotic intelligence, empowering robots to tackle the complexities of the real world.

\begin{table*}[t]
    \centering
    \setcounter{footnote}{1}
    \addtolength{\tabcolsep}{-2pt}
        \begin{tabular}{l l l l c c c c c c l}
            \toprule
            Dataset     & Traj.  & Skill & Scene & \makecell{Detailed \\ Annotation}  &  \makecell{Cam. \\ Calibration}  & Arm Type & \makecell{Dex. \\ Hand} & \makecell{Failure \\ Recovery} & \makecell{Human-in- \\ the-loop} & Collection \\
            \midrule
            RoboNet~\cite{dasari2019robonet}    & 162k      & n/a    & 10      & \xmark & \xmark   & Single & \xmark & \xmark  & \xmark & scripted                                 \\
            BridgeData~\cite{ebert2021bridgedata}  & 7.2k     & 4      & 12      & \xmark & \xmark    & Single & \xmark & \xmark       & \xmark & human teleop                                    \\
            BC-Z~\cite{jang2022bc}       & 26k     & 3      & 1       & \xmark & \xmark   & Single & \xmark & \xmark       & \xmark & human teleop                                    \\
            RT-1~\cite{brohan2022rt1}        & 130k      & 8      & 2       & \xmark & \xmark   & Single & \xmark & \xmark    & \xmark & human teleop                                    \\
            RH20T~\cite{fang2023rh20t}       & 13k      & 33     & 7   & \xmark & \cmark    & Single & \xmark & \xmark  & \xmark & human teleop                                    \\
            RoboSet~\cite{bharadhwaj2024roboagent}     & 98.5k     & 6      & 11      & \xmark & \xmark    & Single & \xmark & \xmark  & \xmark & 30\% human / 70\%  scripted      \\
            BridgeData~V2~\cite{walke2023bridgedata}  & 60.1k  & 13     & 24      & \xmark & \xmark    & Single & \xmark & \xmark  & \xmark & 85\% human / 15\%  scripted \\
            DROID~\cite{khazatsky2024droid} & 76k & 86 & 564 & \xmark & \cmark & Single & \xmark & \xmark & \xmark  & human teleop \\
            RoboMIND~\cite{wu2024robomind} & 55k & 36 & n/a & \xmark & \cmark & Single+Dual & \cmark & \xmark & \xmark & human teleop \\
            \textcolor{gray}{Open X-Embodiment~\cite{padalkar2023open}}   & \textcolor{gray}{1.4M}      & \textcolor{gray}{217}     & \textcolor{gray}{311}   & \textcolor{gray}{(\xmark)} & \textcolor{gray}{\xmark}     & \textcolor{gray}{Single+Dual} & \textcolor{gray}{\xmark} & \textcolor{gray}{\xmark}   & \textcolor{gray}{\xmark} &  \textcolor{gray}{dataset aggregation}  \\
            \midrule
             \textbf{\projname Dataset} & \textbf{1M+} & 87 & 106 & \cmark & \cmark & Dual & \cmark & \cmark & \cmark & human teleop \\
            \bottomrule                                                 \\
        \end{tabular}
    \vspace{-10pt}
    \addtolength{\tabcolsep}{2pt}
    \setcounter{footnote}{0}
    \captionsetup{width=\textwidth}
    \caption{\textbf{Comparison to existing 
    datasets.} \projname features the largest number of trajectories \textit{to date}. 
    We replicate real-world environment at a 1:1 scale for the
    industrial and retail scenarios, which are barely present before. 
    Extensive human annotations are offered, including item, scene, skill (sub-task segmented), and task-level annotations.
    Notably, to expand data applicability and potential,
    we include imperfect data (\textit{i.e.,} failure recovery data with annotated error states) and tasks 
    with dexterous hands.  To ensure data quality,
    we adopt a human-in-the-loop philosophy: the policy learning is performed on collected demonstrations. The deployment results are adopted as feedback to improve the 
    collection protocol. 
    } 
    \vspace{-6pt}
    \label{tab:dataset_comp}
\end{table*}

\textbf{Contribution.} 1) We construct \projname dataset, a multifarious robot learning dataset accompanied by open-source tools to advance research on policy learning at scale. As a pioneering initiative, \projname employs 
an inclusive
optimized pipeline, from scene configuration, task design, data collection, to human-in-the-loop verification, which ensures unparalleled data quality. 2) We propose \modelname, a robot foundation policy using latent action representations to unlock web-scale pre-training on web data. 
Empowered by \projname dataset, it outperforms prior generalist policies in generalization and dexterity. 

\textbf{Limitation.} All 
evaluations are conducted in real-world scenarios. We are currently developing the simulation environment, 
aligning with the real-world setup and aiming to reflect real-world policy deployment outcome. It would thereby facilitate fast and reproducible evaluation.

\section{Related Work}
\textbf{Data scaling in robotics.} Robot learning datasets from automated scripts or human teleoperation have enabled policy learning, with early efforts like RoboTurk~\cite{ajay2018roboturk} and BridgeData~\cite{ebert2021bridgedata} offering small-scale datasets with 2.1k and 7.2k trajectories, respectively. 
Larger datasets, such as RT-1~\cite{brohan2022rt1} (130k trajectories), expand scopes yet remain limited to few environments and skills. 
Open X-Embodiment~\cite{padalkar2023open} aggregates various datasets into a unified format, growing to more than 1.4 million trajectories, as a consequence it suffers from significant variability in embodiments, observation perspectives, and inconsistent data quality, limiting its overall effectiveness. More recently, DROID~\cite{khazatsky2024droid} moves towards scaling up scenes for greater diversity by crowd-sourcing demonstrations yet falls short in data scale and quality control.
Prior datasets above generally face limitations in data scale, task practicality, and scenario naturalness, compounded by inadequate quality assurance and hardware restrictions, which impedes generalist policy training. As shown in \Cref{tab:dataset_comp}, 
our dataset addresses these 
gap adequately.
We build a data collection facility spanning five scenarios to reconstruct real-world diversity and authenticity. With over 1 million trajectories gathered by skilled teleoperators through rigorous verification protocols, \projname utilizes humanoid robots equipped with visuo-tactile sensors and dexterous hands to enable multimodal demonstrations, %
setting it apart from previous efforts. Unlike Pumacay \textit{et al.}~\cite{pumacay2024colosseum}, which serves as a simulation benchmark for evaluating generalization, what we propose is a 
full-stack
platform with data, models, benchmarks, and ecosystem.%

\textbf{Policy learning at scale.} 
Robotic foundation models often co-evolve with the development of dataset scale, equipping robots with escalating general-purpose capabilities through diverse, large-scale training.
Several prior arts use web-scale video only to facilitate policy learning given the limited scale of action-labeled robot datasets~\cite{du2024UniPi,black202susie,bu2024clover}. 
Another line of work lies in the use of large, end-to-end models trained on robot trajectories with robotics data scaling up~\cite{kim2024openvla,brohan2023rt2,brohan2022rt1,team2024octo}. For instance, RDT~\cite{liu2024rdt} employs Diffusion Transformers, initially pre-trained on heterogeneous multirobot datasets and fine-tuned on over 6k dual-arm trajectories, showcasing the benefits of pre-training on diverse sources. $\pi_{0}$~\cite{black2024pi_0} uses a pre-trained VLM backbone and a flow-based action expert, advancing dexterous manipulation for complex tasks like laundry. LAPA~\cite{ye2024lapa} introduces the use of latent actions as pre-training targets; however, its latent planning capability is not preserved for downstream tasks. Building on a variety of innovative ideas from recent research, we advance the field by transferring web-scale knowledge to robotic control through the adaptation of vision-language models (VLMs) with latent actions, leveraging both human videos and robot data for scalable training. Our work demonstrates how the integration of a latent action planner enhances long-horizon task execution and enables more efficient policy learning, significantly improving upon existing generalist policies.

\section{AgiBot World: Platform and Data}

\projname is a 
full-stack
and open-source embodied intelligence ecosystem. Based on the hardware platform developed by us, \embodiment, we construct 
\projname 
--- 
an open-source robot manipulation dataset collected by more than 100 homogeneous robots, providing high-quality data for challenging tasks spanning a wide spectrum of real-life scenarios.  The latest 
version contains 1,001,552 trajectories, with a total duration of 2976.4 hours, covering 217 specific tasks, 87 skills, and 106 scenes. We 
go
beyond basic tabletop tasks such as \textit{pick-and-place} in lab environments; 
instead, concentrate on real-world scenarios involving dual-arm manipulation, dexterous hands, and collaborative tasks. \projname aims to provide 
an inclusive
benchmark to drive the future development of advanced and robust algorithms.

We plan to release \textit{all} resources to enable 
the community
build
upon \projname. \textbf{The dataset is available under the CC BY-NC-SA 4.0 license}, along with the model
checkpoints and code.

\subsection{Hardware: A Versatile Humanoid Robot} 
The hardware platform is the cornerstone of \projname, determining the lower limit of its quality. The standardization of hardware is also the key to streamlining distributed data collection and ensuring reproducible results. We meticulously develop a novel hardware platform for \projname, distinguished by visuo-tactile sensors, durable 6-DoF dexterous hands with humanoid configuration.

As illustrated in \Cref{fig:teaser}, our robotic platform features dual 7-DoF arms, a mobile chassis, and an adjustable waist. The end effectors are modular, allowing for the use of either a standard gripper or a 6-DoF dexterous hand, depending on task requirements. For tasks necessitating tactile feedback, a gripper equipped with visuo-tactile sensors is utilized. The robot is outfitted with eight cameras: an RGB-D camera and three fisheye cameras for the front view, RGB-D or fisheye cameras mounted on each end-effector, and two fisheye cameras positioned at the rear. Image observations and proprioceptive states, including joint and end-effector positions, are recorded at a control frequency of 30 Hz.

We employ two teleoperation systems: VR headset control and whole-body motion capture control. The VR controller maps the hand gesture to  the end-effector translation and rotation, which is subsequently converted to joint angles through inverse kinematics. The thumbsticks and buttons on the controller enable robot base and body movement, while the trigger buttons control end-effector actuation. However, the VR controller restricts the dexterous hand to only a few predefined gestures. To extensively unlock our robot's capabilities, we adapt a motion capture system which records the data of human joints, including the fingers, and maps them to robot posture, enabling more nuanced control, including individual finger movements, torso pose, and head orientation. This system provides posture flexibility and execution precision that are required in achieving more complex manipulation tasks.

\begin{figure}[t]
    \centering 
    \includegraphics[width=0.99\linewidth]{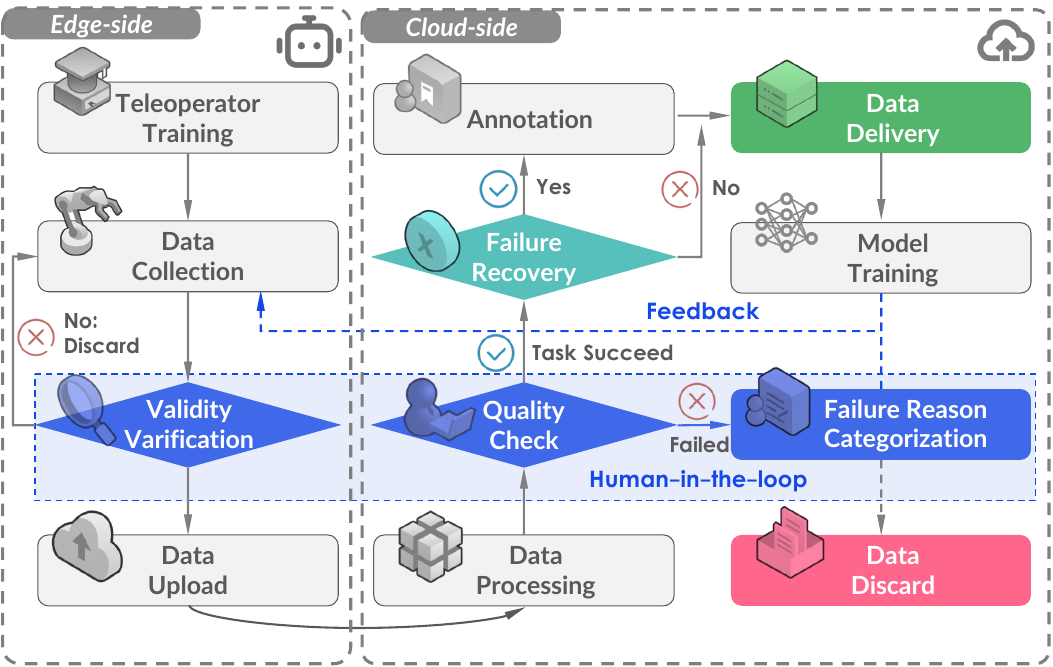}
    \caption{\textbf{Data collection pipeline.} 
    We embrace a human-in-the-loop framework to ensure high quality, enriched with detailed annotations and error recovery behaviors. Human feedback plays a critical role not only in post-collection review but also in actively guiding the data collection process, which is largely overlooked in prior efforts.}
    \vspace{-10pt}
    \label{fig:data_collection}
\end{figure}

\begin{figure*}[t]
    \centering
    \includegraphics[width=0.99\linewidth]{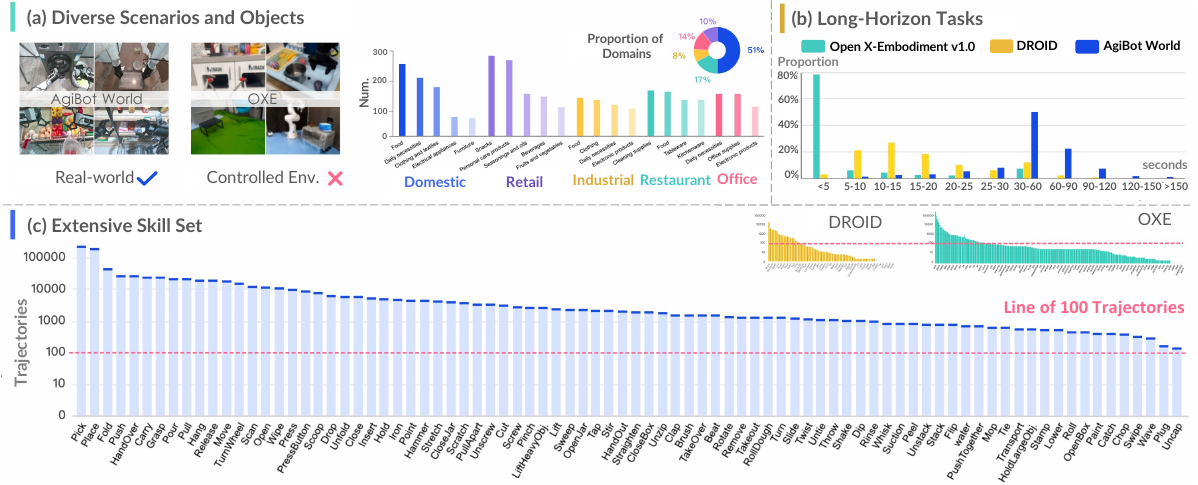}
    \caption{\textbf{Dataset Statistics.}
    \textbf{a)} \projname dataset covers the vast majority of robotic application scenarios, as well as a wide range of interactive objects. \textbf{b)} Our dataset features long-horizon tasks, with the majority of trajectories ranging from 30s to 60s. In contrast, widely used datasets, such as DROID, primarily consist of trajectories ranging from 5s to 20s, while OXE v1.0 predominantly contains trajectories within 5s. \textbf{c)} \projname dataset focuses on 
    valuable atomic skills, spanning a wide spectrum of skills, each supported by a minimum of 100 trajectories (red dashed line above).}
    \vspace{-10pt}
    \label{fig:data_statistics}
\end{figure*}

\subsection{Data Collection: Protocol and Quality}
The data collection session, as shown in~\Cref{fig:data_collection}, can be broadly divided into three phases. (1) Before formally commencing data collection, we first conduct preliminary data acquisition to validate the feasibility of each task and establish corresponding collection standards. (2) After feasibility validation and review of the collection standards, skilled teleoperators arrange the initial scene and formally begin data collection according to the established standards. All data undergoes an initial validity verification locally, such as verifying the absence of missing frames. Once the data is confirmed to be complete, it is uploaded to the cloud for the next phase. (3) During post-processing, the data annotators will verify whether each episode meets the collection standards established in phase 1 and provide language annotations.

\textbf{Failure recovery.}
During data collection, teleoperators may occasionally commit errors, such as inadvertently dropping objects while manipulating the robotic arms. However, they are often able to recover from these errors and successfully complete the task without requiring a full reconfiguration of the setup. Rather than discarding such trajectories, we retain them and manually annotate each with corresponding failure reasons and timestamps. These trajectories, referred to as \textit{failure recovery} data, constitute approximately one percent of the dataset. We consider them invaluable for achieving policy alignment~\cite{zhang2024grape} and failure reflection~\cite{shinn2023reflexion}, essential for advancing the next generation of robot foundation models.

\textbf{Human-in-the-loop.}
Concurrent with feedback collection from data annotators, we adopt a human-in-the-loop approach to assess and refine data quality. This process involves an iterative cycle of collecting a small set of demonstrations, training a policy, and deploying the resulting policy to evaluate data availability. Based on the policy's performance, we iteratively refine the data collection pipeline to address identified gaps or inefficiencies. For instance, during real-world deployment, the model exhibits prolonged pauses at the onset of actions, aligning with data annotator feedback highlighting inconsistent transitions and excessive idle time in the collected data. In response, we revise the data collection protocols and introduce a post-processing step to eliminate idle frames, thereby enhancing the dataset's overall utility for policy learning. This feedback-driven methodology ensures continuous improvement in data quality.

\subsection{Dataset Statistics and Analysis: Beyond Scale}

\projname is developed through a large-scale data collection facility, which spans over 4,000 square meters. This extensive environment contains over 3,000 unique objects in a variety of scenes, meticulously designed to reflect real-world settings. The dataset covers a wide range of scenarios and scene setups, ensuring both scale and diversity in the pursuit of generalizable robot policy.

\textbf{Reconstructing the diversity of the real world.} Key statistics of our dataset are presented in \cref{fig:data_statistics}. \projname provides extensive coverage across five key domains: domestic, retail, industrial, restaurant, and office environments. Within each domain, we further define specific scene categories. For instance, the domestic domain includes detailed environments such as bedrooms, kitchens, living rooms, and balconies, while the retail domain features distinct areas like shelving units and fresh produce sections. Our dataset also features over 3,000 distinct objects, systematically categorized across various scenes. These objects span a wide range of everyday items, including food, furniture, clothing, electronic devices, and more. The distribution of object categories, as illustrated in \Cref{fig:data_statistics}(a), highlights the relative frequency of different object types within each scene.

\textbf{Long-horizon manipulation.}
A distinguishing feature of the \projname dataset is its emphasis on long-horizon manipulation. As shown in \Cref{fig:data_statistics}(b), prior datasets predominantly focus on tasks involving single atomic skills, with most trajectories lasting no more than 5 seconds. In contrast, \projname is built upon continuous and complete tasks composed by multiple atomic skills, like ``make a coffee''. Trajectories in our dataset typically span approximately 30 seconds, some of which last over 2 minutes. We also provide key-frame and instruction annotation for each sub-step to facilitate policy learning in such challenging scenarios.

\textbf{Comprehensive skill coverage.} In terms of task design, while generic atomic skills, such as ``pick-and-place'', dominate the majority of tasks, we have intentionally incorporated tasks that emphasize less frequently used but highly valuable skills, such as ``\texttt{chop}'' and ``\texttt{plug}'' (as shown in \Cref{fig:data_statistics}(c)). This ensures that our dataset adequately represents a broad spectrum of skills, providing sufficient data for each to support robust policy learning.
\begin{figure*}[t]
    \centering
    \includegraphics[width=0.99\linewidth]{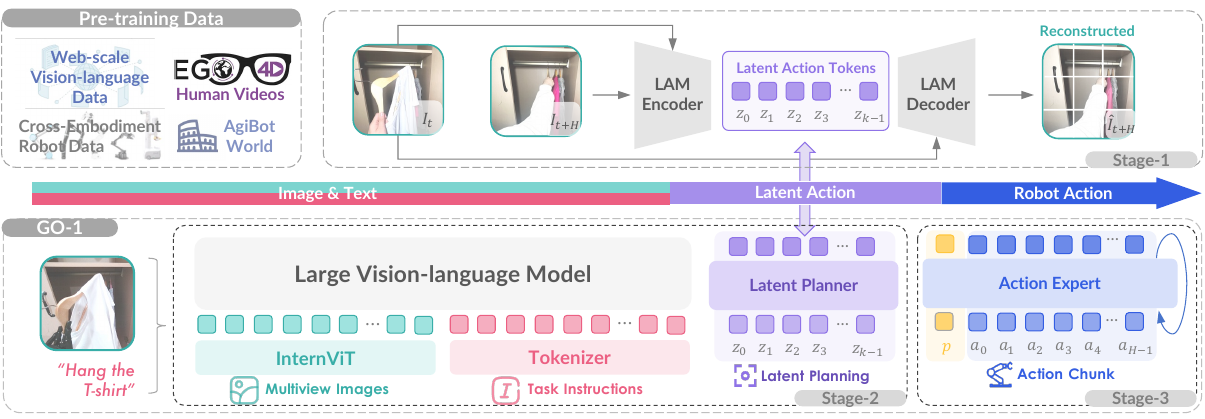}
    \caption{\textbf{We propose \modelname}, a 
    generalist policy
    featuring general reasoning and long-horizon planning capabilities.  The latent action model (LAM) learns universal action representations from web-scale video data~(\textit{i.e.,} human videos from Ego4D), and quantizes them into discrete latent action tokens. The latent planner conducts temporal reasoning through latent action prediction, bridging the gap between image-text inputs and robot actions generated by the action expert.}
    \vspace{-6pt}
    \label{fig:policy_architecture}
\end{figure*}

\section{AgiBot World: Model}

To effectively utilize our high-quality \projname dataset and enhance the policy's generalizability, we propose a hierarchical \textbf{Vi}sion-\textbf{L}anguage-\textbf{L}atent-\textbf{A}ction (ViLLA) framework with three training stages, as depicted in~\Cref{fig:policy_architecture}. Compared to Vision-Language-Action (VLA) model where action is vision-language conditioned, the ViLLA model predicts latent action tokens, 
conditioned on the generation of subsequent robot control actions.

In Stage 1, we project consecutive images into a latent action space by training an encoder-decoder latent action model (LAM) on internet-scale heterogeneous data. This allows the latent action to serve as an intermediate representation, bridging the gap between general image-text inputs and robotic actions. In Stage 2, these latent actions act as pseudo-labels for the latent planner, facilitating embodiment-agnostic long-horizon planning and leveraging the generalizability of the pre-trained VLM. Finally, in Stage 3, we introduce the action expert and jointly train it with the latent planner to support the learning of dexterous manipulation.

\subsection{Latent Action Model}

Despite considerable advancements in gathering diverse robot demonstrations, the volume of action-labeled robot data remains limited relative to web-scale datasets. To broaden the data pool by incorporating internet-scale human videos lacking action labels and cross-embodiment robot data, we employ latent actions~\cite{bruce2024genie} in Stage 1 to model the inverse dynamics of consecutive frames. This approach enables the transfer of real-world dynamics from heterogeneous data sources into universal manipulation knowledge.

To extract latent actions from video frames $\{I_t, I_{t+H}\}$, the latent action model is constructed around an inverse dynamics model-based encoder $\mathbf{I}(z_{t} | I_{t}, I_{t+H})$ and a forward dynamics model-based decoder $\mathbf{F}(I_{t+H} | I_{t}, z_{t})$. The encoder employs a spatial-temporal transformer~\cite{xu2020st-trans} with casual temporal masks following Bruce \textit{et al.}~\cite{bruce2024genie}, while the decoder is a spatial transformer that takes the initial frame and discretized latent action tokens $z_{t} = [z_t^{0}, ..., z_{t}^{k-1}]$ as input, with $k$ set to 4. The latent action tokens are quantized using a VQ-VAE objective~\cite{van2017neural}, with a codebook of size $|C|$.

\subsection{Latent Planner}

With the aim of establishing a solid foundation for scene and object understanding and general reasoning ability,
the ViLLA model harnesses a VLM pre-trained on web-scale vision-language data and incorporates a latent planner for embodiment-agnostic planning within the latent action space. We use InternVL2.5-2B~\cite{chen2025expandingperformanceboundariesopensource} as the VLM backbone due to its strong transfer learning capabilities. The two-billion parameter scale has proven effective for robotic tasks in our preliminary experiments, as well as in prior studies~\cite{liu2024rdt,black2024pi_0}. Multiview image observations are first encoded using InternViT before being projected into the language space. The latent planner consists of 24 transformer layers, which enable layer-by-layer conditioning from the VLM backbone with full bidirectional attention. 

Specifically, given multiview input images $\left(I_{t}^{h}, I_{t}^{l}, I_{t}^{r}\right)$ (typically from the head, left wrist, and right wrist) at timestep $t$, along with a language instruction $l$ describing the ongoing task, the latent planner predicts latent action tokens: $\mathbf{P}\left(z_{t}|I_{t}^{h}, I_{t}^{l}, I_{t}^{r}, l\right)$, with supervision produced by the LAM encoder based on the head view: $z_{t} := \mathbf{I}(I_{t}^{h}, I_{t+H}^{h})$.
Since the latent action space is orders of magnitude smaller than the discretized low-level actions used in OpenVLA~\cite{kim2024openvla}, this approach also facilitates the efficient adaptation of general-purpose VLMs into robot policies.

\subsection{Action Expert}

To achieve high-frequency and dexterous manipulation, Stage 3 integrates an action expert that utilizes a diffusion objective to model the continuous distribution of low-level actions~\cite{bu2024robodual}. Although the action expert shares the same architectural framework as the latent planner, their objectives diverge: the latent planner generates discretized latent action tokens through masked language modeling, while the action expert regresses low-level actions via an iterative denoising process. Both expert modules are conditioned hierarchically on preceding modules, including the action expert itself, ensuring coherent integration and information flow within the dual-expert system.

The action expert decodes low-level action chunks, denoted by ${A}_t = [{a}_t, {a}_{t+1}, ..., {a}_{t+H}]$ with $H=30$, using proprioceptive state $p_t$ over an interval of $H$ timesteps: $\mathbf{A}\left(A_t|I_{t}^{h}, I_{t}^{l}, I_{t}^{r}, p_t, l\right)$. During inference, the VLM, latent planner, and action expert are synergistically combined within the generalist policy \modelname, which initially predicts $k$ latent action tokens and subsequently conditions the denoising process to produce the final control signals.

\begin{figure*}[t]
    \centering
    \includegraphics[width=\linewidth]{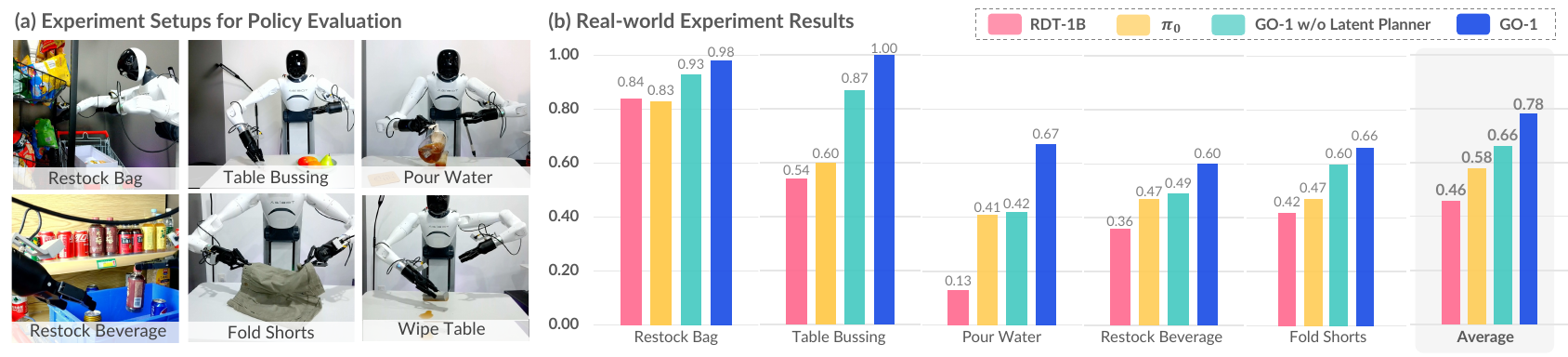}
    \caption{\textbf{Is \modelname a more powerful robot generalist policy?} We evaluate \modelname against previous generalist policy RDT-1B and our baseline 
    without the latent planner, with all policies pre-trained on \projname \texttt{beta}.
    Across all tasks and comparisons, \modelname outperforms baselines by a large margin. The incorporation of latent planner 
    boosts 
    performance on complex tasks such as ``Fold Shorts'' and improves generalizability in task ``Restock Beverage'' in great extent.}
    \vspace{-6pt}
    \label{fig:model_comparison}
\end{figure*}

\begin{figure}[t]
    \centering
    \includegraphics[width=0.99\linewidth]{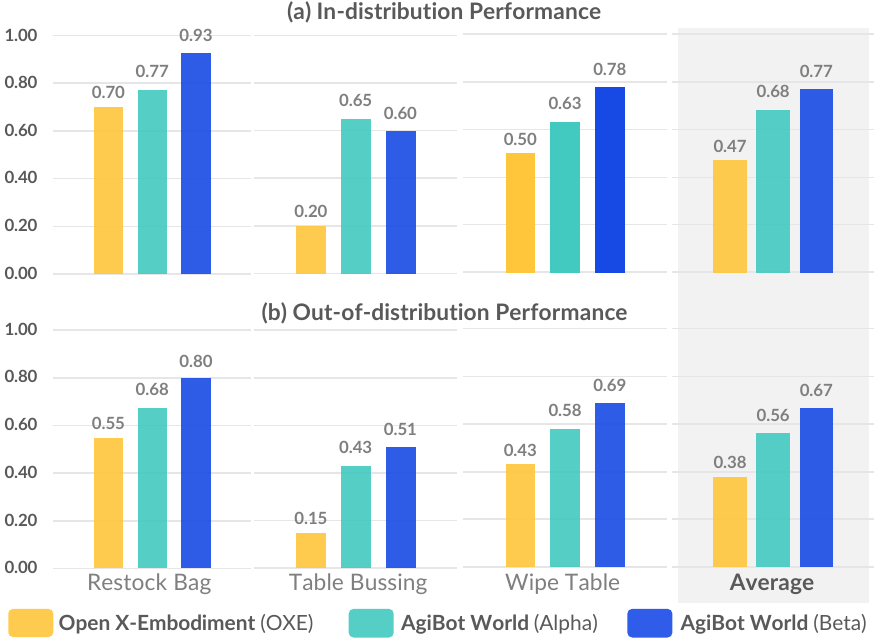}
    \caption{\textbf{Does \projname dataset improve policy performance and generalizability?} Policies pre-trained on our dataset outperform those trained on OXE in both seen (0.77 \textit{v.s.} 0.47) and out-of-distribution scenarios (0.67 \textit{v.s.} 0.38).
    }
    \vspace{-10pt}
    \label{fig:data_comparison}
\end{figure}

\section{Experiment and Analysis}\label{sec:experiment}

We evaluate the real-world performance of policies pre-trained on different data sources including the \projname dataset, demonstrating the effectiveness
credited from 
the \modelname model in policy learning.

\subsection{Experiment Setup}

\subsubsection{Evaluation Tasks} 

Here we choose a comprehensive set of tasks that span various dimensions of policy capabilities from \projname for evaluation, including \textbf{tool-usage} (Wipe Table), \textbf{deformable objects manipulation} (Fold Shorts), \textbf{human-robot interaction} (Handover Bottle), \textbf{language-following} (Restock Beverage), etc. Moreover, we design 2 unseen scenarios for each task, covering position generalization, visual distractors, and language generalization, delivering thorough generalization evaluations for policies. The evaluated tasks, also partially shown in~\Cref{fig:model_comparison}, are:
1) ``Restock Bag'': Pick up the snack from the cart and place it on the supermarket shelf;
2) ``Table Bussing'': Clear tabletop debris into the trash can; %
3) ``Pour Water'': Grasp the kettle handle, lift the kettle and pour water into the cup; %
4) ``Restock Beverage'': Pick up the bottled beverage from the cart and place it on the supermarket shelf; %
5) ``Fold Shorts'': Fold the shorts laid flat on the table in half twice;
6) ``Wipe Table'': Clean water spills using the sponge. %

\noindent
\textbf{Scoring rubrics.} The evaluation metric employs a normalized score, computed as the average across 10 rollouts per task, scenario, and method. Each episode scores 1.0 for full success, with fractional scores for partial success, enabling a nuanced performance assessment.

\subsubsection{Implementation Details} 

The \projname \texttt{alpha} dataset is an early-stage subset, containing partial tasks and roughly 14\% of the trajectories in the full \texttt{beta} version. (\textit{a.k.a.} last row in~\Cref{tab:dataset_comp}). Following the completion of the third-stage pre-training, the pre-trained \modelname exhibits basic competency in task completion. Unless otherwise specified, we further enhance the model by fine-tuning it using high-quality, task-specific demonstrations, enabling adaptation to new tasks for evaluation. For \modelname, fine-tuning is conducted with a learning rate of 2e-5, a batch size of 768, and 30,000 optimization steps. 

\begin{figure}[t]
    \centering
    \includegraphics[width=0.99\linewidth]{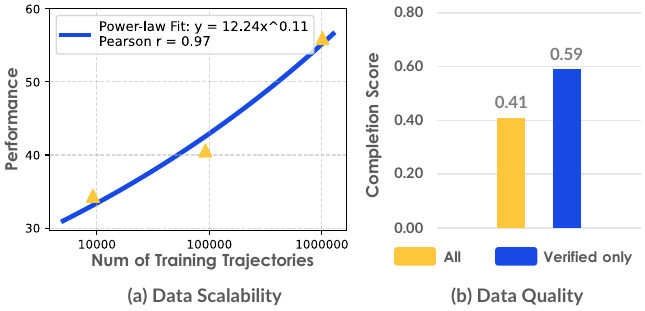}
    \caption{\textbf{Further analysis on}: a) how model performance scales with data size, and b) the impact of filtering undesirable data through manual review on policy learning.} 
    \vspace{-10pt}
    \label{fig:analysis}
\end{figure}

\subsection{Does \projname boost policy learning at scale?}

We choose the open-source RDT~\cite{liu2024rdt} model to study how much the \projname dataset can help policy learning. 
The task completion scores for three tasks are detailed in~\Cref{fig:data_comparison}. Models pre-trained on the \projname dataset demonstrate a significant improvement in the ``Table Bussing'' task, nearly tripling performance. On average, the completion score increases by 0.30 and 0.29 for in-distribution and out-of-distribution setups, respectively. Notably, the \projname \texttt{alpha} dataset, despite having a significantly smaller data volume than OXE (\textit{e.g.}, 236h compared to $\sim$2000h), achieves a higher success rate, underscoring the exceptional data quality of our dataset.

\subsection{Is \modelname a more capable generalist policy?}
We evaluate \modelname on five tasks of varying complexity, categorized by their visual richness and task horizon. The results, as shown in~\Cref{fig:model_comparison}, are averaged over 30 trials per task, with 10 trials conducted in a seen setup and 20 trials under variations or distractions. \modelname significantly outperforms RDT and $\pi_{0}$, particularly in tasks such as ``Pour Water'', which demands robustness to object positions, and ``Restock Beverage'', which highlights instruction-following capabilities. The inclusion of the latent planner yields an average improvement of 0.12 task completion score.

\subsection{Does GO-1's ability scale with data size?}
To investigate whether a power-law scaling relationship exists between the size of pre-training data and policy capability, we conduct an analysis using 10\% subsets of the \texttt{alpha}, 100\% \texttt{alpha}, and \texttt{beta} dataset, where the number of training trajectories are ranged from 9.2k to 1M. We evaluate the out-of-the-box performance of resulting policies on four seen tasks in pre-training. As shown in \cref{fig:analysis}(a), the policy's performance exhibits a predictable power-law scaling relationship with the number of trajectories, supported by a Pearson correlation coefficient of $r=0.97$.

\subsection{How does data quality impact policy learning?}
We explore the impact of quality checks introduced in our human-in-the-loop data collection on policy learning. Specifically, we provide an ablation study by fine-tuning an RDT model using both verified (528 trajectories) and unverified (482 trajectories) data from the ``Wipe Table'' task. Verification refers to our ``human-in-the-loop'' quality assurance method. As shown in \cref{fig:analysis}(b), being larger in quantity does not necessarily translate to improved performance, while a smaller set of human-verified data yields a 0.18 boost in the completion score, underscoring the importance of high-quality data for policy learning.

\section{Conclusion}

We introduce \projname, an open-source ecosystem aimed at democratizing access to large-scale, high-quality robot learning datasets. It is complete with toolchains and foundation models to advance embodied general intelligence through community collaboration. Our dataset distinguishes itself through unparalleled scale, diversity, and quality, underpinned by carefully crafted tasks. Policy learning evaluations confirm \projname's value in enhancing performance and generalizability. To further explore its impact, we develop \modelname, a generalist policy utilizing latent actions for web-scale pre-training. \modelname excels in real-world complex tasks, outperforming existing generalist policies and demonstrating scalable performance with increased data volume. We invite the broader community to collaborate in fostering an ecosystem and maximizing the potential of our extensive dataset.

\bibliographystyle{IEEEtran.bst}
\bibliography{bib_acronym, bib_local, bib_global}

\bigskip
\appendix

\section*{Acknowledgement}
We thank Remi Cadene and the LeRobot community for their support and collaboration. In addition, we are grateful to Shu Jiang, Chengshi Shi, Shenyuan Gao, Yixuan Pan, Yi Xin and Peng Gao for their 
fruitful discussions. We also extend our gratitude to the open-source community for 
the insightful feedback and engagement throughout this project.

\section*{Contributions}

\subsection*{\textbf{\color{yellow}Core Contributors}}
\noindent
\textit{\color{deemph}The whole span of project, including data collection, algorithm, experiment, and writing}

\noindent
Qingwen Bu, 
Guanghui Ren, 
Chiming Liu, 
Modi Shi, \\
Chengen Xie, 
Xindong He, 
Jianheng Song, \\
Yuxiang Lu, 
Siyuan Feng

\medskip
\subsection*{\color{cvprblue}\textbf{Algorithm}}

\noindent
\textit{\color{deemph}Technical roadmap, model training, and evaluation}

\noindent
\textbf{\color{cvprblue}Roadmap and Methodology}

\noindent
Yao Mu, Li Chen, Yan Ding 

\noindent
\textbf{\color{cvprblue}Pre-training}

\noindent
Yi Liu, Yuxin Jiang, Xiuqi Cui

\noindent
\textbf{\color{cvprblue}Post-training}

\noindent
Ziyu Xiong, Xu Huang, Dafeng Wei

\noindent
\textbf{\color{cvprblue}Deployment \& Evaluation}

\noindent
Guo Xu

\subsection*{\color{cvprblue}\textbf{Product \& Ecosystem}}
\noindent
\textit{\color{deemph}System architecture design, project management, community engagement}

\noindent
Chengyue Zhao,
Shukai Yang, 
Huijie Wang, 
Yongjian Shen, \\
Jialu Li, 
Jiaqi Zhao, 
Jianchao Zhu, 
Jiaqi Shan

\subsection*{\color{cvprblue}\textbf{Manuscript Preparation}}
\noindent
\textit{\color{deemph}Manuscript outline, writing, and revising}

\noindent
Jisong Cai, Chonghao Sima 

\medskip
\subsection*{\textbf{\color{cvprblue}Data Curation}}
\noindent
\textit{\color{deemph}Data collection, quality check}

\noindent
Cheng Ruan, Jia Zeng, Lei Yang

\medskip
\subsection*{\color{cvprblue}\textbf{Hardware \& Software Development}}
\noindent
\textit{\color{deemph}Hardware design, embedded software development}

\noindent
Yuehan Niu, Cheng Jing, Mingkang Shi, Chi Zhang, \\
Qinglin Zhang, Cunbiao Yang, Wenhao Wang,
Xuan Hu

\medskip
\subsection*{\color{DeepPurple}\textbf{Project Co-lead and Advising}}
\noindent
\textit{\color{deemph}Research direction, project coordination, technical advising}

\noindent
Maoqing Yao, \textcolor{blue}{\texttt{yaomaoqing@agibot.com}} \\
Yu Qiao, \textcolor{blue}{\texttt{qiaoyu@pjlab.org.cn}}  \\ 
Hongyang Li, \textcolor{blue}{\texttt{hongyang@sii.edu.cn}}  \\ 
Jianlan Luo, 
Bin Zhao, 
Junchi Yan,
Ping Luo

\section*{Change Log}

\subsection*{\textbf{Jan 2025}: agibot-world alpha version release}
\begin{itemize}
    \item A sub-split (around 10\%) of the dataset, containing 92,214 trajectories.
\end{itemize}

\subsection*{\textbf{March 2025}: full data release and technical report}
\begin{itemize}
    \item Initial submission to arXiv and reserach blog
    \item This version includes the complete manuscript with all sections: introduction, methodology, results, discussion, and appendix.
\end{itemize}

\subsection*{\textbf{July 2025}: IROS camera-ready update}
\begin{itemize}
    \item We conduct additional comparisons with $\pi_{0}$.
\end{itemize}

\end{document}